\documentclass[10pt,twocolumn,letterpaper]{article}

\usepackage{cvpr}              
\usepackage{array}
\usepackage{graphicx}
\usepackage{amsmath}
\usepackage{amssymb}
\usepackage{booktabs}
\usepackage[dvipsnames]{xcolor}

\definecolor{cvprblue}{rgb}{0.21,0.49,0.74}
\usepackage[pagebackref,breaklinks,colorlinks,citecolor=cvprblue]{hyperref}

    \def\confName{CVPR}
    \def\confYear{2024}

\title{Long-MIL: Scaling Long Contextual Multiple Instance Learning for Histopathology Whole Slide Image Analysis}

\author{Honglin Li$^{1,2}$, Yunlong Zhang$^{1,2}$, Chenglu Zhu$^{2}$,  Jiatong Cai$^{2}$, Sunyi Zheng$^{2}$, Lin Yang$^{2\thanks{Corresponding author.}}$ \\
$^{1}$Zhejiang University, $^{2}$Westlake University \\
\tt\small {honglin\_li@zju.edu.cn,yanglin@westlake.edu.cn}
}

\begin{document}
\maketitle

\begin{abstract}
Histopathology image analysis is the golden standard of clinical diagnosis for Cancers. In doctors daily routine and computer-aided diagnosis, the Whole Slide Image (WSI) of histopathology tissue is used for analysis. 
Because of the extremely large scale of resolution, previous methods generally divide the WSI into a large number of patches, then aggregate all patches within a WSI by Multi-Instance Learning (MIL) to make the slide-level prediction when developing computer-aided diagnosis tools. 
However, most previous WSI-MIL models using global-attention without pairwise interaction and any positional information, or self-attention with absolute position embedding can not well handle shape varying large WSIs, e.g. testing WSIs after model deployment may be larger than training WSIs, since the model development set is always limited due to the difficulty of histopathology WSIs collection.
To deal with the problem, in this paper, we propose to amend position embedding for shape varying long-contextual WSI by introducing Linear Bias into Attention, and adapt it from 1-d long sequence into 2-d long-contextual WSI which helps model extrapolate position embedding to unseen or under-fitted positions. 
We further utilize Flash-Attention module to tackle the computational complexity of Transformer, which also keep full self-attention performance compared to previous attention approximation work. 
Our method, Long-contextual MIL (Long-MIL) are evaluated on extensive experiments including 4 dataset including WSI classification and survival prediction tasks to validate the superiority on shape varying WSIs. 
The source code will be open-accessed soon.
\end{abstract}    
\begin{figure}[htbp]
  \centering
    \begin{subfigure}{0.48\textwidth}
        \centering
        \includegraphics[width=\textwidth]{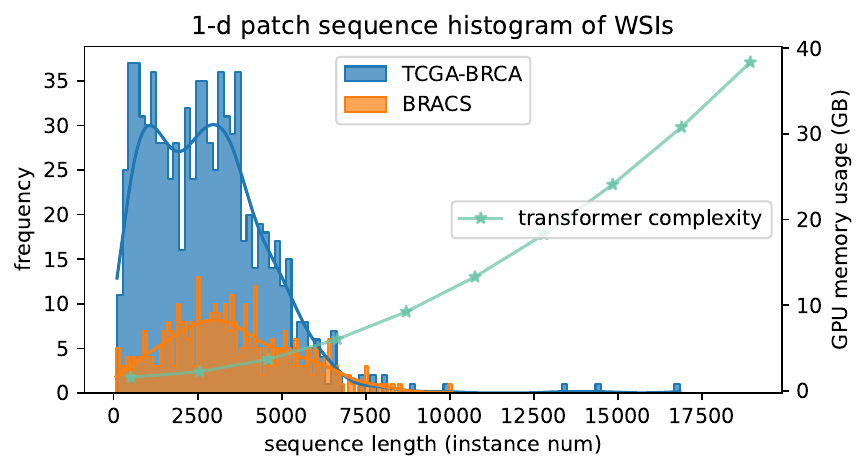}
    \end{subfigure}
    \hfill
    \begin{subfigure}{0.48\textwidth}
        \centering
        \includegraphics[width=0.8\textwidth]{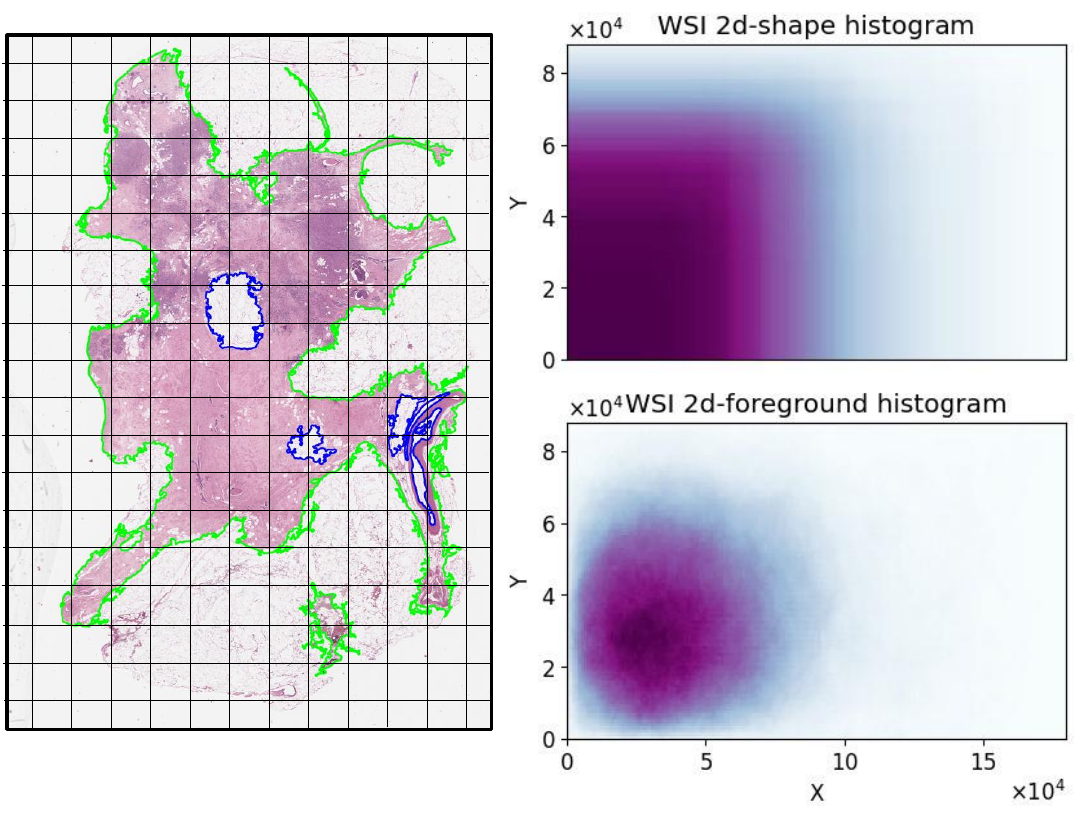}
    \end{subfigure}
   
   \caption{Two attributes of WSIs' shape pose challenges for contextual modelling. 1) extremely long sequence even in $20 \times$ magnitude, which could be quadruple in $40 \times$. 2) shape variance of WSIs and their foreground distribution.}
   \label{figure1}
   \vspace*{-5 mm}
\end{figure}

\section{Introduction}
\label{sec:intro}
Though digital pathology images have been widely used for Cancer diagnosis \cite{lu2021data, NEURIPS2021_10c272d0, Zhang2022DTFDMILDF,zhang2022benchmarking,cai2021generalizing,zhang2022weakly,shui2023deformable} and prognosis \cite{chen2022scaling, Chen_2021_ICCV} via automatic computer-assisted analysis, the Giga-pixels of resolution, as large as $150,000 \times 150,000$ pixels\cite{lu2021data, Chen_2021_ICCV}, still poses great challenges on both precise annotations and efficient computation for model training \cite{Li_2023_CVPR}.
Thus, previous methods \cite{chen2022scaling, Chen_2021_ICCV,Li_2023_CVPR,bulten2022artificial,li2021dual,ACMIL} focus on developing annotation- \& computational- efficient learning to cope with those problems by employing Multiple Instance Learning (MIL) \cite{maron1997framework,pmlr-v80-ilse18a} with only WSI-level supervision.
The MIL is defined as predicting the highest level category of instances as final result within a bag, where all small patches in the WSI are regarded as instances to constitute a bag sample \cite{campanella2019clinical,lu2021data, NEURIPS2021_10c272d0,ACMIL}, and the category of WSI corresponds to the max lesion level of all patch instances.

Currently, there are mainly three steps (or mainstream genre) for WSI-MIL analysis framework: 1) accessing better instance-level patch embedding via Self-supervised Learning \cite{mae,dino,chen2022scaling,li2021dual}. 2) designing WSI head architectures \cite{lu2021data, NEURIPS2021_10c272d0, Zhang2022DTFDMILDF} and train the head with frozen instance embedding. 3) fine-tuning embedding and WSI head simultaneously \cite{zhang2023promptmil, Li_2023_CVPR} with top-K instances for better task-specific results. 
Here in this paper, we focus on the step-2 and find that there are still some room for improvement: 
Firstly, the global-attention used in AB-MIL, DS-MIL, CLAM, etc. \cite{pmlr-v80-ilse18a,li2021dual,lu2021data,Zhang2022DTFDMILDF} with light computational cost (compared to self-attention) can not model contextual information within WSI (including local-spatial context and long-range dependency). In other words, the relation between different instances, or pairwise interaction is ignored, which is quite useful indeed for prediction decision making \cite{chen2021slide, NEURIPS2021_10c272d0} and should be performed by self-attention.
Secondly, though the self-attention computation complexity on long sequence WSI instances (Figure \ref{figure1}a) can be alleviated by Linear Attention\cite{xiong2021nystromformer,wang2020linformer}, also used in TransMIL \cite{xiong2021nystromformer,NEURIPS2021_10c272d0}, its softmax approximation only get sub-optimal performance compared to self-attention as pointed out by Tri et.al \cite{dao2022flashattention}. 
Most importantly, shape varying large WSIs (as shown in Figure \ref{figure1}b) makes absolute position embedding for WSI-MIL used in \cite{chen2022scaling,NEURIPS2021_10c272d0} can not be well generalized (see more visual illustration Figure \ref{figure2}).

Above issues present a strong need for better positional embedding with input length extrapolation ability as well as memory-efficient Transformer for shape varying, long contextual WSIs modelling. 
Motivated by recent advancements of Large Language Model \cite{openai2023gpt,touvron2023llama,iyer2022opt,zhang2022opt} on long-context modelling \cite{su2021roformer,press2021alibi,chi2022kerple,chen2023longlora,chi2022receptive}, we propose to leverage relative positional embeddings \cite{su2021roformer,press2021alibi,chi2022kerple} to replace traditional absolute embeddings\cite{vaswani2017attention,Kan2021vision}.
Specifically, we employ Attention with Linear Bias (ALiBi) \cite{press2021alibi} that biases query-key attention scores with a penalty which is proportional to their distance. 
Since such Bias is original designed as linearly dependent on words index distance for 1-d language modelling, we adapt it as also linearly dependent on the Euclidean distance for the 2-d large scale WSI. 
In addition to the positional embedding, we further use FlashAttention (FA) \cite{dao2022flashattention} for long sequence memory efficient Transformer modelling especially on memory saving, which also keeps full ability like original self-attention, compared to Linear Attention.
Assisted by the efficient Transformer and relative 2-d spatial positional embedding provided by FA and ALiBi respectively, we can model both the semantic and spatial-positional correlation among instances within the extremely long sequence of 2-d shape varying WSI. 
Our main contributions can be concluded into 3 folds: 
\begin{itemize}
    \item [1)] We propose to adapt Attention with Linear Bias into 2-d positional embedding for shape varying large WSI, which provides input length extrapolation ability to generalize on different input size and under-fitted positions of WSI during testing.
    \item [2)] We use FlashAttention for efficient Transformer computation to replace current self-attention and Linear Attention, which helps us modelling long sequence of WSI instances in lightweight GPU memory and computational cost without no information loss. 
    \item [3)] Our WSI-analysis experiments are performed on both diagnosis and prognosis task on 4 WSI dataset including Breast, Stomach, Colon and Rectal Carcinoma, which show strong universality of the method and practical potential for real-world applications.
\end{itemize}

\section{Related Work}
\subsection{Multiple Instance Learning for WSI Analysis.}
Whole Slide Images (WSIs) contain a rich set of visual information that can aid in pathological diagnosis. \cite{campanella2019clinical,lu2021data} However, accurate annotation of cell-level information within WSIs is labor-intensive and time-consuming.\cite{campanella2019clinical,lu2021data,chen2022scaling,NIPS2017_3f5ee243}. 
To address this issue, weakly-supervised methods have gained popularity in pathology WSI analysis. Attention-based Multi-Instance Learning (AB-MIL) \cite{pmlr-v80-ilse18a} is adopted to learns instance weights adaptively, allowing the model to focus on informative regions within the WSIs. 
This approach significantly reduces the annotation burden on pathologists while still providing valuable insights for patient-level diagnosis. 
In the context of weakly-supervised pathology WSI analysis, several innovative approaches, DS-MIL, CLAM, DTFD, etc. \cite{pmlr-v80-ilse18a,li2021dual,lu2021data,Zhang2022DTFDMILDF} have been proposed. However, their utilized global-attention with light computational cost can not model WSI contextual information, which is used in pathologist diagnosis decision making \cite{chen2021slide, NEURIPS2021_10c272d0}. The fine-grained details and global contextual information can also be captured by multi-scale modeling \cite{chen2022scaling,li2021dual}. Graph Network  \cite{chen2021slide,li2018graph,guan2022node,chan2023histopathology} is also useful to make model be context-aware. Similar to this, HIPT \cite{chen2022scaling} and TransMIL \cite{NEURIPS2021_10c272d0} explored the advantages of Transformer with pairwise interaction learning ability to model such contextual information. Since Transformer can be generalized to Graph Network \cite{dwivedi2020generalization}, both modelling the pairwise interaction, thus in this paper we focus more on Transformer and try to adapt it better to fit the shape varying and long context properties of WSI.  
\begin{figure*}[htbp]
  \centering
   \includegraphics[width=0.9\linewidth]{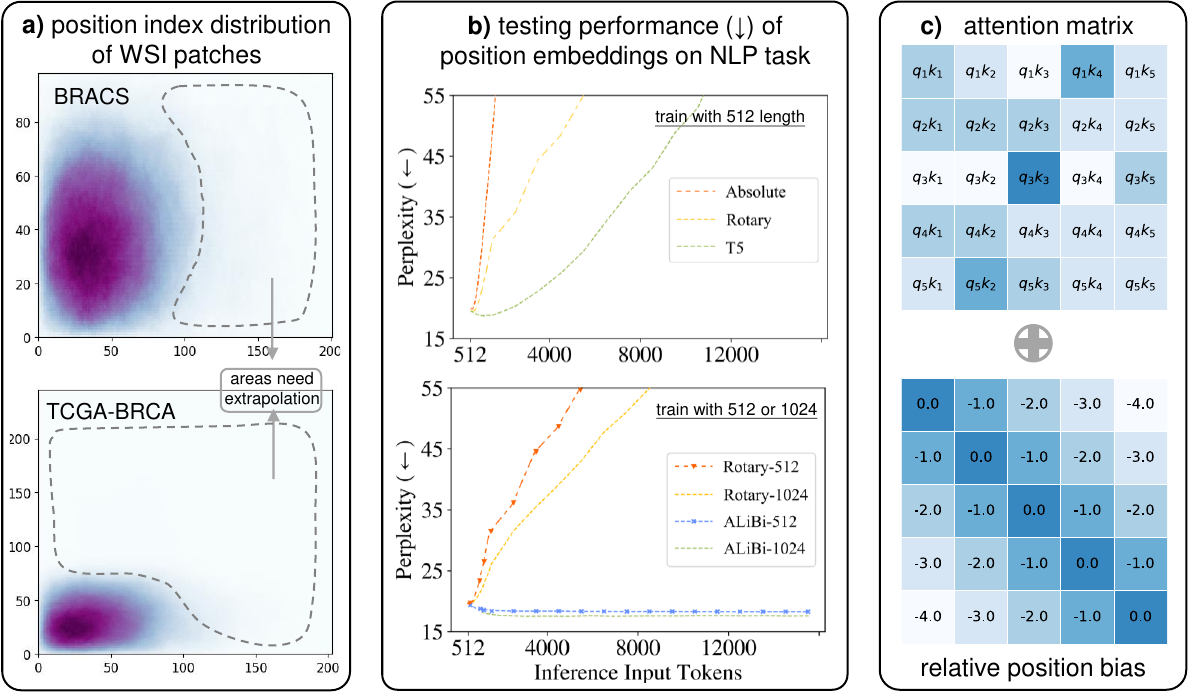}
   \caption{Here we show why the shape varying WSI need long sequence positional embeddings with extrapolation ability. \underline{\textbf{a)}} The normalized  2-d position index distribution of WSI foreground patches mainly scattered within a circle (index$<$100). Thus the positions in area enclosed by the dashed line suffers under-fitting if using traditional positional embedding as shown in \underline{\textbf{b)}}, where performance gets quickly decreased during testing on unseen longer input length (fail to extrapolate). Though longer training input can smooth the problem in NLP, it is unable to replicate this to histopathology due to the scarcity of WSI training data. Thus to address this issue, ALiBi is a more appropriate tool with strong extrapolation ability, whose intuitive visualization can be find in \underline{\textbf{c)}}: longer distance needs larger penalty or bias to attention score. Since the relative position bias is pre-defined and needs no training, it shows strong generalization on unseen long positions.}
   \label{figure2}
\vspace*{-5 mm}
\end{figure*}

\subsection{Efficient Transformer for Long Range Arena} 
The primary goal of this area is to alleviate the computation and memory complexity of self-attention mechanism on long sequence input. 
Earliest modifications simply sparsify the attention matrix, including Blockwise\cite{qiu2019blockwise}, Local Attention\cite{parmar2018image}, Sparse Transformer\cite{child2019generating} and Longformer\cite{beltagy2020longformer} with large or dilated stride.
Extend to above fixed patterns, some work \cite{vyas2020fast, wang2020cluster,roy2020efficient,tay2020sparse,kitaev2020reformer} using learnable patterns in a data-driven fashion, e.g. Reformer\cite{kitaev2020reformer} introduces a hash-based similarity measure to efficiently cluster tokens into chunks. 
Linformer\cite{wang2020linformer} technique leverage low-rank approximations of the self-attention matrix, decomposing the $N \times N$ matrix to $N \times k$.
The kernels also serve as a approximation of the attention matrix, including Performers\cite{katharopoulos2020Transformers}, Linear Transformers \cite{choromanski2020masked} and Random Feature Attention (\cite{peng2021random})
Another popular method of reducing computation cost is to reduce the resolution of the sequence, hence reducing computation cost by a commensurate factor, e.g. Perceiver\cite{jaegle2021perceiver}, Swin Transformer\cite{liu2021swin}. The recent Nystr\"{o}mformer\cite{xiong2021nystromformer}, been used in TransMIL\cite{NEURIPS2021_10c272d0} of WSI-MIL, can be seem like kernel-based low-rank approach. Above work mainly focus on a \textit{light approximation of self-attention or using sparse attention, which is indeed worse than the full attention} \cite{dao2022flashattention}.

Another lines of work try to merge RNN and Transformer, e.g. Transformer-XL\cite{dai2019Transformer} proposed a segment-level recurrence mechanism that connects multiple segments and blocks, and now is widely used in most successful LLM\cite{openai2023gpt,zhang2022opt,touvron2023llama}.
Attention Free Transformer \cite{zhai2021attention} replaces dot-product self-attention with a computationally efficient alternative. RWKV \cite{peng2023rwkv}, Linear Recurrent Units \cite{orvieto2023resurrecting}, State space models \cite{gu2022efficientlys4} and its variants \cite{dao2022hungry,poli2023hyena} are also proposed, but \textit{these recurrent ability of attention is designed for text sequence with causal or auto-regressive property, not fit well for image recognition}. 
Recent work like FlashAttention \cite{dao2022flashattention} and others \cite{rabe2022selfattention, jang2019mnnfast} using chunked computation scheme and IO-aware mechanism to \textit{be memory-efficient and gain full ability like self-attention}. We argue that this kind of work is more suitable for WSI analysis task since the total sequence length of most WSI-MIL tasks will be around 10-20k, but self-attention approximation or attention free work try to scaling the model into infinite length for language modelling, which will lose some self-attention ability.

\subsection{Long Contextual Positional Embedding} 
Recently, explore positional embeddings to longer sequences play a vital role in LLM to solving long context modelling \cite{dao2022flashattention, wang2020linformer}. Initially, absolute positional embedding assign a positional vector and adds it to the embedding vector by the first work predefined sinusoidal function\cite{vaswani2017attention}. Followed by the success of BERT\cite{devlin2018bert}, learnable absolute positional embeddings have been applied to the task of masked language modeling\cite{devlin2018bert,liu2019roberta,Clark2020ELECTRA,Lan2020albert}, Autoregressive-decoding \cite{radford2018improving,radford2019language}, and sequence-to-sequence\cite{Gehring2017seq2seq,lewis2019bart} settings. 

A line of work studies the ways to extrapolate sinusoidal positional embeddings to longer sequences by randomly shifting absolute positions during training \cite{kiyono2021shiftedAbs} or augmenting with continuous signals \cite{Likhomanenko2021CAPE}.
While being easy to implement, \textit{it is challenging to extend absolute positional embeddings to unseen longer sequence lengths}, which is known as the length extrapolation issue\cite{press2021alibi}.
As opposed to the modeling of absolute position, relative positional embeddings (RPE) that model the positional difference has become popular in the literature \cite{shaw2018rpe,huang2018music,dai2019Transformer,yang2019xlnet,huang2020rpe,he2021deberta,ke2021rethinking,chen2021simple}. 
In particular, the T5 model that considers bucketed relative distances and log-binning has been shown to perform well on various Transformer architectures\cite{raffel2019exploring}. 
Rotary positional embedding (RoPE) \cite{su2021roformer} encodes the position with rotations, with the rotation's property, the query-key product exhibits a positional difference. However, \textit{the RoPE only fits well trained length, showing poor performance on unseen or seldom seen length or position}. Attention with Linear Bias (ALiBi) \cite{press2021alibi,chi2022kerple} proposes to add relative positional bias term directly to the attention matrix, which provide the \textit{extrapolation ability ( training short but testing long)} for language models.
Despite above work focuses on the NLP domain, recent work has applied positional embeddings to other domains such as vision \cite{Kan2021vision} and speech \cite{Likhomanenko2021CAPE}. 
However, similar problem also happens in ViT's \cite{Kan2021vision} naive 2-d absolute position embedding, making it not fit our WSI task well.
Since there is no special long sequence modelling problem is most computer vision tasks, the histopathology WSI analysis tasks present us a special challenging for 2-d image long-sequence modelling to tackle with in this paper.

\section{Method}
\subsection{Attention-based WSI Analysis}
Given a WSI $X$ as input, the goal is to make slide-level prediction $\hat{Y}$ by learning a classifier $f(X;\theta)$.
X is firstly patched into a long sequence of small instances $X=\{x_1,..., x_N\}$ because of its extremely large resolution, where N is the number of instance.
The slide-level supervision $Y$ is given by doctors who considering the latent label $y_i$ of all instance $x_i$. Most previous work \cite{campanella2019clinical,lu2021data,chen2022scaling,NIPS2017_3f5ee243} try to model this process by a Max-pooling operation, so initially, this annotation process can be treated as:
  \begin{equation}
    Y=\max\{y_1,..., y_N\}.
    \label{EQ1}
  \end{equation}
Since the end-to-end training from raw image input to WSI-level output is impossible because of large memory usage, conventional approaches convert it into two separate stages: 
Firstly, convert all small patches into instance embeddings $Z=\{z_1,...,z_N\}$ by a pretrained backbone such as CNN\cite{he2016deep} or ViT\cite{Kan2021vision}, which refers to general features from public ImageNet, or learned on the related dataset to extract the domain-specific representations\cite{chen2022scaling,Kang_2023_CVPR}. Then, aggregate all patches' features within a slide and producing the slide-level prediction $Y = g(Z; \theta)$. In this paper, we mainly focus on the latter one, where $g$ is an global-attention function followed by a linear classifier head as:
  \begin{equation}
      Y = \sigma (\sum_{i=1}^N a_i z_i),
      \label{EQ2}
  \end{equation}
where $a_i$ is attention weights and $\sigma(\cdot)$ is a linear head.

However, above vanilla method utilize global-attention (assign adaptive weight to each instance to make simple summation or pooling) can not model the interactions among different instances. Thus, to handle this problem, Transformer with self-attention is employed in TransMIL\cite{NEURIPS2021_10c272d0} and HIPT\cite{chen2022scaling}, 
where the attention sublayer computes the attention scores for the $i$-th query ${q}_i \in {R}^{1\times d}$, ($1 \leq i \leq N$) in each head, where $d$ is the head dimension.
In other words, each instance will compute a attention score as interaction with all instances.
These attention scores are then multiplied by the values to return the output of the attention sub-layer as:
  \begin{equation}
       o_i = \text{softmax} ( q_i  K^\top)  v_i,
      \label{EQ_SA}
  \end{equation}
where the $ \{Q, K, V \} \in {R}^{N \times d}$ is got by linear transform from $Z$ with different parameters and $O \in {R}^{N \times d}$ is the output attention score. Given $O$, which encodes the interactions among instances (or pairwise interactions), we can further use Equation \eqref{EQ2} and input $O$ to replace $Z$ for final prediction, also mean-pooling and class token in ViT \cite{Kan2021vision} can be adopted. Note that here we omit dropout, FFN, residual connection and some detailed blocks in Transformer for simplicity.

\subsection{Memory-efficient Attention for WSI}
Though above Transformer with self-attention can well model the interactions among different instances, its computational memory usage is too heavy $O(N^2)$ for long sequence of WSI (average 8k tokens in $20\times$ magnitude of 224x224 patch size) due to the interactive attention score calculation (as shown in Figure \ref{figure4} ablations).
Also find details in supplementary materials about the computational complexity and GPU memory usage during forward and backward of self-attention or linear-attention.
Instead of using attention matrix approximation by Nystr\"{o}mformer\cite{xiong2021nystromformer} in TransMIL\cite{NEURIPS2021_10c272d0}, we seek FlashAttention (FA) for help without information lose but show comparable speed.
We omit the algorithm of FA here for its method complexity and interactions with hardware, please check the original paper and supplementary materials for details.

\begin{figure*}[htbp]
  \centering
   \includegraphics[width=1.0\linewidth]{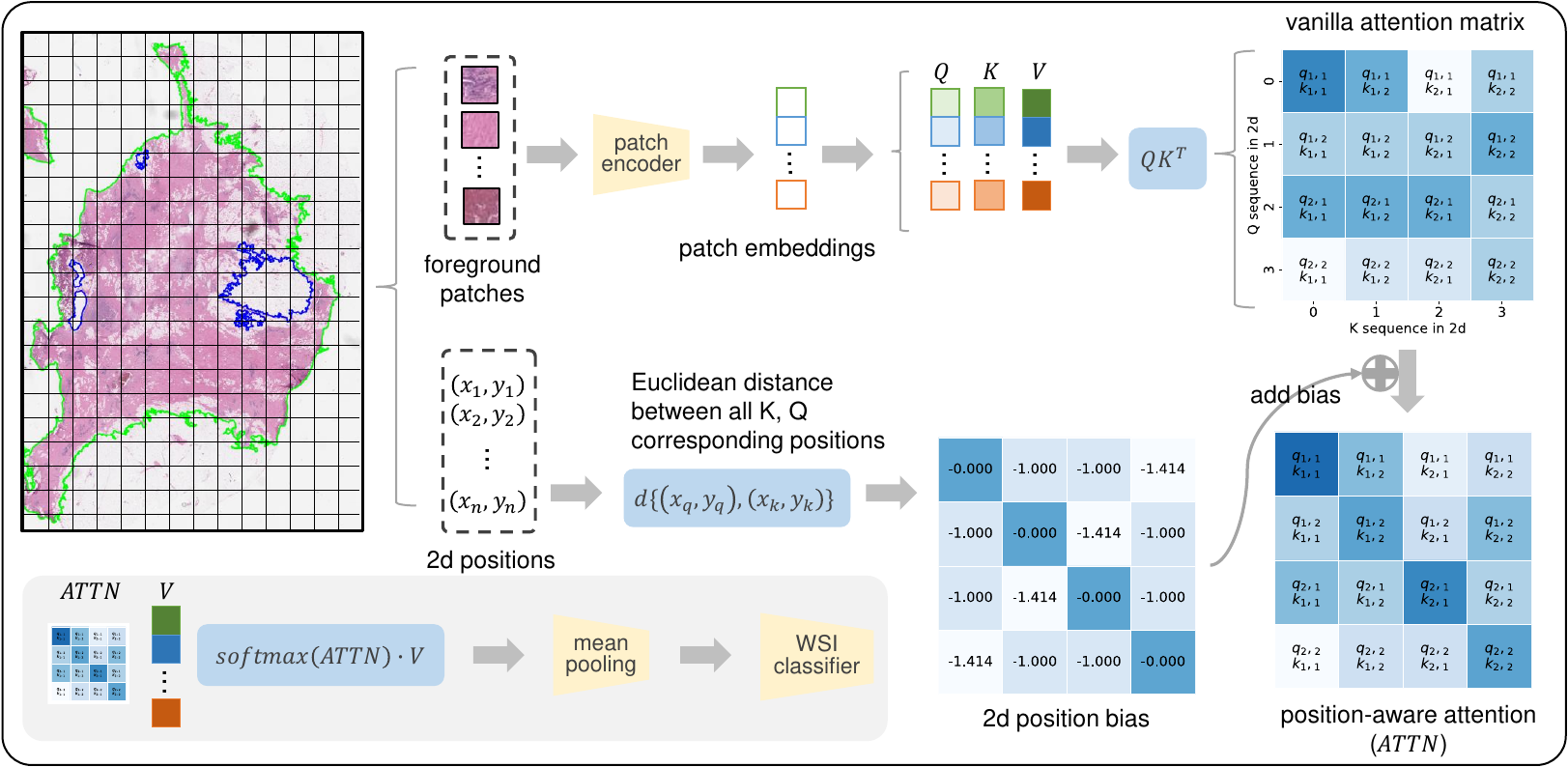}
   \caption{Long-MIL framework for WSI spatial contextual information fusion. 1) Preparing foreground patch feature embedding with its $Q,K,V$ transformation and 2-d positions of WSIs. 2)  Performing pairwise computations among all positions within a WSI to get distances as 2-d positional bias matrix for attention. 3) Calculating the vanilla attention matrix and add it with above 2-d position bias matrix, then using the standard Transformer process to finish WSI prediction.}
   \label{figure3}
   \vspace*{-5 mm}
\end{figure*}

\subsection{Attention with Positional Bias}

Since the operation in Equation \eqref{EQ_SA} is position-agnostic, Transformer \cite{vaswani2017attention} try to model contextual interactions by incorporate position information, which can also be seen as inductive bias for self-attention. \textbf{Absolute} positional embeddings assign a positional vector $p_m$ to each position $m$ and adds it to the embedding vector as: $z_i = z_i + p_{m,i}$.

To improve long sequence ability, \textbf{Relative} positional embeddings that model the positional difference $m-n$ has become popular. Rotary positional embedding (RoPE) \cite{su2021roformer} encodes the position with rotations: $f(q_m,m)=R_m  q_m$ where $R_m$ is a rotation matrix with angles proportional to $m$. With the rotation's property, the query-key product exhibits a positional difference: 
\begin{equation}
f(q_m,m) f(k_n,n) ^\top =  q_m R_{n-m} k_n ^\top.
\label{eq_rope}
\end{equation}
The core idea of RoPE is to insert position $m, n$ signal on $q, k$ and reflect the relative position on the newly attention matrix, where the rotary can match this product property well and result in above rotation matrix. Though the RoPE is design for 1-d language sequence, it can also be extended to 2-d paradigm for application on WSI analysis. We omit it here, please check Supplementary Material for details.

However, RoPE need to be well trained or fine-tuned on unseen or seldom seen long length \cite{liu2023scaling,chen2023longlora,xiao2023efficient}, as shown in Figure \ref{figure2}b. So here, we introduce \textbf{Attention with Linear Bias} \cite{press2021alibi} to the 2-d shape varying WSI analysis (2d-ALiBi).
The main modification is to add bias term after the query-key dot producted attention matrix. For the original 1-d ALiBi \cite{press2021alibi}, the bias is a static, non-learned matrix $\text{softmax}({q}_m {K}^\top + \tau \cdot [-(m-1), ..., -2, -1, 0]),$ computed by the distance between tokens from different position (closer position tokens using smaller bias):
\begin{equation}
    {q}_m {k}_n^\top - \tau \left| m-n \right| 
\label{eq_alibi_1d}
\end{equation}
where scalar $\tau$ is multi-head coefficients fixed before training.
With this predefined distance aware or position-aware bias matrix (see Figure \ref{figure2}c for visualization), no matter how long the unseen sequence is, the relative positions can always be well encoded, or in other word, extrapolation. This property fit well for our shape varying WSI analysis, where too many masks will be add to seldom seen positions, resulting sub-optimal learning of RoPE. 

To extend the ALiBi for 2-d WSI, the bias matrix in Equation \eqref{eq_alibi_1d} can be convert by calculating the 2-d Euclidean distance among positions (as shown in Figure \ref{figure3} for a simple 2-d $(2\times 2) \cdot (2\times 2)$ position matrix for visualization, we give a larger matrix in supplementary materials):
\begin{equation}
    {q}_{m} {k}_{n}^\top - \tau \sqrt{ \left| m_{j}-n_{j} \right|^2 +  \left| m_{k}-n_{k} \right|^2},
\label{eq_alibi_2d}
\end{equation}
where the $j,k$ represents the 2-d Coordinate axes.

\begin{table*}[htbp]
  \begin{center}   
  \begin{tabular}{m{3.9cm}<{}||m{2.2cm}<{\centering}m{2.2cm}<{\centering}|m{2.2cm}<{\centering} m{2.2cm}<{\centering}}
  
    \midrule[1.2pt]
    & \multicolumn{4}{c}{BRACS tumor subtyping}\\
    \cline{2-5}
    
      & \multicolumn{2}{c|}{\underline{ViT-S \cite{Kang_2023_CVPR}}} & \multicolumn{2}{c}{\underline{ViT-S DINO (our pretrain)}}  \\
    Method & F1 & AUC & F1 & AUC\\
  \midrule
  KNN (Mean) & 0.503\footnotesize{$\pm$}0.011 & 0.691\footnotesize{$\pm$}0.007 & 0.430\footnotesize{$\pm$}0.029 & 0.649\footnotesize{$\pm$}0.008 \\
  KNN (Max) & 0.472\footnotesize{$\pm$}0.009 & 0.771\footnotesize{$\pm$}0.018 & 0.416\footnotesize{$\pm$}0.019 & 0.645\footnotesize{$\pm$}0.007  \\
    Mean-pooling &  0.534\footnotesize{$\pm$}0.026 & 0.741\footnotesize{$\pm$}0.017  &  0.487\footnotesize{$\pm$}0.034 & 0.717\footnotesize{$\pm$}0.020  \\
  Max-pooling &  0.649\footnotesize{$\pm$}0.032 & 0.843\footnotesize{$\pm$}0.018  &  0.598\footnotesize{$\pm$}0.032 & 0.818\footnotesize{$\pm$}0.006   \\
  
  AB-MIL \cite{pmlr-v80-ilse18a} &  0.668\footnotesize{$\pm$}0.032 & 0.866\footnotesize{$\pm$}0.016  & 0.621\footnotesize{$\pm$}0.048 & 0.837\footnotesize{$\pm$}0.035  \\
  DS-MIL\cite{li2021dual} &  0.607\footnotesize{$\pm$}0.044 & 0.824\footnotesize{$\pm$}0.028  &  0.622\footnotesize{$\pm$}0.063 & 0.808\footnotesize{$\pm$}0.033  \\
  CLAM-SB \cite{lu2021data} & 0.647\footnotesize{$\pm$}0.020 & 0.836\footnotesize{$\pm$}0.021 &  0.627\footnotesize{$\pm$}0.032 & 0.836\footnotesize{$\pm$}0.009 \\
  DTFD-MIL MaxS\cite{Zhang2022DTFDMILDF} &  0.597\footnotesize{$\pm$}0.025 & 0.874\footnotesize{$\pm$}0.026  & 0.521\footnotesize{$\pm$}0.059 & 0.807\footnotesize{$\pm$}0.016 \\
  DTFD-MIL AFS\cite{Zhang2022DTFDMILDF} &  0.608\footnotesize{$\pm$}0.083 & 0.869\footnotesize{$\pm$}0.018  & 0.538\footnotesize{$\pm$}0.053 & 0.824\footnotesize{$\pm$}0.011 \\
  \midrule[0.5pt]
  TransMIL \cite{NEURIPS2021_10c272d0} &  0.648\footnotesize{$\pm$}0.054 & 0.835\footnotesize{$\pm$}0.031 & 0.591\footnotesize{$\pm$}0.049 & 0.798\footnotesize{$\pm$}0.029  \\
  FlashAttention (FA) \cite{dao2022flashattention} &  \underline{0.675\footnotesize{$\pm$}0.028} & \textbf{0.882\footnotesize{$\pm$}0.010}  & {0.655\footnotesize{$\pm$}0.042} & {0.843\footnotesize{$\pm$}0.013} \\
   FA + 2d-RoPE (ours) &  {0.673\footnotesize{$\pm$}0.028} & {0.875\footnotesize{$\pm$}0.015}  & \underline{0.663\footnotesize{$\pm$}0.009} & \underline{0.847\footnotesize{$\pm$}0.005}  \\
    FA + 2d-ALiBi (ours) &  \textbf{0.714\footnotesize{$\pm$}0.038} & \underline{0.881\footnotesize{$\pm$}0.013}  & \textbf{0.674\footnotesize{$\pm$}0.041} & \textbf{0.864\footnotesize{$\pm$}0.010}  \\
  \midrule[1.2pt]
  \end{tabular}   
  \caption{
    \textbf{Slide-Level Tumor Subtyping} on BRACS by using two pre-trained embeddings.
\textbf{Top Rows.} Various WSI-MIL architectures with global-attention (no interaction among different instances).
\textbf{Bottom Rows.} Previous state-of-the-art model TransMIL (using Linear self-attention and learnable absolute position embedding), and our proposed FlashAttention and relative positional embedding modules.} 
  \label{T1} 
  \end{center}   
  \vspace*{-5 mm}
  \end{table*}

\subsection{Long-MIL framework and implementation}
To realize long contextual MIL modelling and better WSI analysis performance, the overall framework (as depicted in Figure \ref{figure3}) of our method includes 3 stages:
\begin{itemize}
\item [1)] Segmenting and patching WSI into instances, then save its corresponding foreground patch feature embedding and 2-d positions for preparation.
\item [2)] Performing pairwise computations among all positions within a WSI to get distances as 2-d positional bias matrix for attention. For efficiency, we pre-compute a large matrix $(300\times300) \cdot (300\times300)$, then for each WSI, using the foreground standardized positions as indexing to get sub-matrix for needing (2-d ALiBi). 
\item [3)] Calculating the vanilla attention matrix, then add it with above 2-d position bias matrix, thus the 2-d position-aware attention (ATTN) is obtained. Using the newly ATTN to make $softmax$ operation to get pairwise attention score and finally we finish the long contextual spatial information interaction and fusion by the position aware attention score adaptive summation. This step is fully supported and accelerated by FlashAttention\cite{dao2022flashattention} ( inputting the bias matrix as the mask term, which is widely used in NLP for causal generation).
\end{itemize}
\begin{table*}[htbp]
  \begin{center}   
  \begin{tabular}{m{3.9cm}<{}||m{2.2cm}<{\centering}m{2.2cm}<{\centering}|m{2.2cm}<{\centering} m{2.2cm}<{\centering}}
  
    \midrule[1.2pt]
    & \multicolumn{4}{c}{TCGA-BRCA tumor subtyping}\\
    \cline{2-5}
    
      & \multicolumn{2}{c|}{\underline{ViT-S DINO (our pretrain)}} & \multicolumn{2}{c}{\underline{ResNet-50 (ImageNet pretrain)}}  \\
    Method & F1 & AUC & F1 & AUC\\
  \midrule
  KNN (Mean) & 0.671\footnotesize{$\pm$}0.055 & 0.843\footnotesize{$\pm$}0.020 & 0.585\footnotesize{$\pm$}0.048 & 0.742\footnotesize{$\pm$}0.016 \\
  KNN (Max) & 0.652\footnotesize{$\pm$}0.038 & 0.718\footnotesize{$\pm$}0.004 & 0.516\footnotesize{$\pm$}0.033 & 0.691\footnotesize{$\pm$}0.016  \\
    Mean-pooling &  0.832\footnotesize{$\pm$}0.042 & 0.936\footnotesize{$\pm$}0.010  &  0.751\footnotesize{$\pm$}0.049 & 0.861\footnotesize{$\pm$}0.026  \\
  Max-pooling &  0.843\footnotesize{$\pm$}0.020 & 0.935\footnotesize{$\pm$}0.008  &  0.780\footnotesize{$\pm$}0.027 & 0.886\footnotesize{$\pm$}0.301   \\
  AB-MIL \cite{pmlr-v80-ilse18a} &  0.854\footnotesize{$\pm$}0.013 & 0.940\footnotesize{$\pm$}0.015  & 0.760\footnotesize{$\pm$}0.046 & 0.851\footnotesize{$\pm$}0.057  \\
  DS-MIL\cite{li2021dual} &  0.850\footnotesize{$\pm$}0.053 & 0.933\footnotesize{$\pm$}0.011  &  \underline{0.797\footnotesize{$\pm$}0.036} & 0.894\footnotesize{$\pm$}0.029  \\
  CLAM-SB \cite{lu2021data} & 0.853\footnotesize{$\pm$}0.020 & 0.926\footnotesize{$\pm$}0.021 &  0.779\footnotesize{$\pm$}0.035 & 0.878\footnotesize{$\pm$}0.027 \\
  DTFD-MIL MaxS\cite{Zhang2022DTFDMILDF} &  0.799\footnotesize{$\pm$}0.056 & 0.900\footnotesize{$\pm$}0.035  & 0.653\footnotesize{$\pm$}0.604 & 0.798\footnotesize{$\pm$}0.236 \\
  DTFD-MIL AFS\cite{Zhang2022DTFDMILDF} &  0.841\footnotesize{$\pm$}0.025 & 0.921\footnotesize{$\pm$}0.012  & 0.787\footnotesize{$\pm$}0.037 & 0.897\footnotesize{$\pm$}0.027 \\
  \midrule[0.5pt]
  TransMIL \cite{NEURIPS2021_10c272d0} &  0.831\footnotesize{$\pm$}0.037 & 0.928\footnotesize{$\pm$}0.015 & 0.741\footnotesize{$\pm$}0.126 & 0.854\footnotesize{$\pm$}0.051  \\
  FlashAttention (FA) \cite{dao2022flashattention} &  {0.861\footnotesize{$\pm$}0.035} & \underline{0.943\footnotesize{$\pm$}0.010}  & 
  \textbf{0.800\footnotesize{$\pm$}0.014} & {0.901\footnotesize{$\pm$}0.014} \\
   FA + 2d-RoPE (ours) &  \underline{0.863\footnotesize{$\pm$}0.018} & {0.939\footnotesize{$\pm$}0.030}  & {0.772\footnotesize{$\pm$}0.065} & \underline{0.907\footnotesize{$\pm$}0.017}  \\
    FA + 2d-ALiBi (ours) &  \textbf{0.871\footnotesize{$\pm$}0.040} & \textbf{0.946\footnotesize{$\pm$}0.011}  & {0.781\footnotesize{$\pm$}0.047} & \textbf{0.919\footnotesize{$\pm$}0.008}  \\
  \midrule[1.2pt]
  \end{tabular}   
  \caption{
    \textbf{Slide-Level Tumor Subtyping on TCGA-BRCA} by using two pre-trained embeddings. We show various WSI-MIL architectures with global-attention, Linear self-attention, learnable absolute position embedding and our proposed FlashAttention with relative positional embedding modules.} 
  \vspace*{-5 mm}
  \label{T2} 
  \end{center}   
  \end{table*}
\section{Experiments}
  
In this section, we present the performance of the proposed method and compare it with various baselines. Ablation experiments are performed to further study the proposed method and for paper length, more experimental results are presented in the Supplementary.

\noindent \textbf{Datasets and Tasks.}

We use four datasets to evaluate our method.
For the slide-level \textbf{tumor subtyping} performance, our method is evaluated on two datasets: 

BReAst Carcinoma Subtyping (BRACS) \cite{brancati2021bracs} collect H$\&$E stained Histology Images, containing 547 WSIs for three lesion types, i.e., benign, malignant and atypical, which are further subtyped into seven categories. Here, since the WSIs number is limited, we only perform three class subtyping. The WSIs are segmented in $20 \times$ magnitude and non-overlapping patching with $224\times 224$ size.
The Cancer Genome Atlas Breast Cancer (TCGA-BRCA) \cite{tomczak2015review, petrick2021spie} is a public dataset for breast invasive carcinoma cohort for Invasive Ductal Carcinoma versus Invasive Lobular Carcinoma subtyping. The WSIs are segmented into non-overlapping tissue-containing patches with a size of $256 \times 256$ (keep consistency to previous work \cite{chen2022scaling}) at $20\times$ magnification patches were curated from 1038 WSIs.
For the slide-level \textbf{survival prediction}, we includes 2 TCGA histology datasets:  1) A combination dataset of the Colon adenocarcinoma and Rectum adenocarcinoma Esophageal carcinoma (TCGA-COADREAD), which includes 316 WSIs as used in HIPT \cite{chen2022scaling}. 2) Stomach adenocarcinoma (TCGA-STAD) dataset including 321 WSIs.


\noindent \textbf{Pretraining Backbones.}

Our work mainly focus on the WSI-head results based on some good pretrained embeddings for histopathology\cite{chen2022scaling, Kang_2023_CVPR}. 
For tumor subtyping of BRACS, we utilize the embedding proposed in \cite{Kang_2023_CVPR} which add some pathology-domain specific augmentations to DINO\cite{dino} pretraining process on their collected diverse histology patches. We also compare the results on ResNet-50\cite{he2016deep} embedding pretrained ImageNet just like most previous work do\cite{li2021dual,NEURIPS2021_10c272d0,lu2021data}. The multiple kinds of embedding backbone also help demonstrating the consistency of the method. Thus for TCGA-BRCA, we also pretrain the backbone with DINO\cite{dino} by extracting all patch raw image.
For fair comparisons on the experiments of TCGA-related data of survival prediction with previous work, we use the pretrained VIT-small embedding proposed by HIPT\cite{chen2022scaling} which is finished with DINO\cite{dino} on about 10k TCGA histopathology WSI data. We omit the ResNet-50 embedding for survival prediction since it get quite low unacceptable results.

\noindent \textbf{Implementation Details.}

We train our model with PyTorch on a RTX-3090 GPU, with a WSI-level batchsize of 1, learning rate of 1e-4, and weight decay of 1e-2. To save memory usage and boosting the self-attention operation, we employ Flash-Attention \cite{dao2022flashattention} instead. The position bias term is fed into flash-attention on the traditional mask term. Also checking our codes in supplementary material for further details.

\subsection{Slide-level Tumor Subtyping}
\noindent \textbf{Evaluation Metrics.}
For all the experiments, we report the macro-AUC and macro-F1 scores since all these dataset suffering class imbalance.
For TCGA-BRCA, we perform 10-fold cross-validation with the same data split adopted in HIPT \cite{chen2022scaling}.
Besides, the dataset BRACS is officially split into training, validation and testing, thus the experiment is conducted 5-times with different random seeds.
The mean and standard variance values of performance metrics are reported for multi-runs or cross-validation runs.

\noindent \textbf{Baselines for comparison.}
We first show the results of Mean-/Max- pooling and KNN for traditional evaluation. Then we directly evaluate several classical WSI-MIL methods, including AB-MIL\cite{pmlr-v80-ilse18a}, DS-MIL\cite{li2021dual}, CLAM \cite{lu2021data}, DTFD-MIL \cite{Zhang2022DTFDMILDF}. Then we compare our method with some state-of-the-art combining position embedding on Transformer, TransMIL\cite{NEURIPS2021_10c272d0}. We omit HIPT \cite{chen2022scaling} in this task since it need WSI larger than a threshold.
We finally show our proposed long-contextual position embedding module, including RoPE and ALiBi in 2-d form. 

\noindent \textbf{Results Analysis:}
For \underline{BRACS} 3-categories tumor subtyping, the results are reported in Table \ref{T1}. We can first observe that both FA and 2-d positional embedding show their improvement respectively. For FA, attributing to its full self-attention for pairwise interaction ability, it shows better performance compared to all global-attention modules \cite{pmlr-v80-ilse18a,li2021dual,lu2021data,Zhang2022DTFDMILDF} and especially TransMIL \cite{NEURIPS2021_10c272d0} which use linear attention approximation. 
We also notice that backbone embedding extracted from ViT-S pretrained by Kang et al. \cite{Kang_2023_CVPR} showing superiority compared our DINO pretrained model on the training set of BRACS data, which may because of its large dataset learned generalization. The embedding of ResNet-50 pretrained on ImageNet is included in supplementary materials. 
Based on better embedding with \cite{Kang_2023_CVPR}, the AUC score only shows slight improvement equipped with 2-d positional embedding, especially for RoPE without extrapolation ability showing no improvement also in F1-score. However, our 2d-ALiBi still show significant improvement on F1-score, which is quite important for multi-class problem.

For \underline{TCGA-BRCA} 2-categories tumor subtyping, the results are reported in Table \ref{T2}. For fair comparisons to former work on this data, we utilize ViT-S pretrained in HIPT\cite{chen2022scaling}, as well as commonly used ResNet-50 pretrained on ImageNet. We find that there is limited improvement of our method on ResNet-50 (right column of Table \ref{T2}) given that its knowledge or semantic domain gap to histopathology data. In other words, such kind of embedding may lose the spatial information after layers of convolution thus result in bad performance. With better embedding pretrained on pathology domain (left column of Table \ref{T2}), we find that our method show promising improvement. Thus we argue that only a good pretrained embedding helps discovering the potential of positional embedding.

\begin{table}[htbp]
  \begin{center}   
  \begin{tabular}{m{2.8cm}<{}||m{2.2cm}<{\centering}m{2.2cm}<{\centering}}
    \midrule[1.2pt]
    Method & COADREAD & STAD\\
  \midrule
  AB-MIL \cite{pmlr-v80-ilse18a} &  0.566\footnotesize{$\pm$}0.075 & 0.562\footnotesize{$\pm$}0.049  \\
  AMISL \cite{yao2020whole} &  0.561\footnotesize{$\pm$}0.088 & 0.563\footnotesize{$\pm$}0.067  \\
  DS-MIL\cite{li2021dual} &  0.470\footnotesize{$\pm$}0.053 & 0.546\footnotesize{$\pm$}0.047  \\
  GCN-MIL \cite{li2018graph} & 0.538\footnotesize{$\pm$}0.049 & 0.513\footnotesize{$\pm$}0.069 \\

  HIPT \cite{chen2022scaling} &  0.608\footnotesize{$\pm$}0.088 & 0.570\footnotesize{$\pm$}0.081 \\

  TransMIL \cite{NEURIPS2021_10c272d0} &  0.597\footnotesize{$\pm$}0.134 & 0.564\footnotesize{$\pm$}0.080 \\
  FlashAttention(FA)  & {0.603\footnotesize{$\pm$}0.048} & {0.568\footnotesize{$\pm$}0.074}  \\
   FA + 2d-RoPE &   \underline{0.613\footnotesize{$\pm$}0.077} &  \underline{0.575\footnotesize{$\pm$}0.045}  \\
    FA + 2d-ALiBi &  \textbf{0.624\footnotesize{$\pm$}0.057} & \textbf{0.589\footnotesize{$\pm$}0.066} \\

  \midrule[1.2pt]
  \end{tabular}   
  \caption{
    \textbf{Slide-Level Survival Prediction} based on HIPT\cite{chen2022scaling} pre-trained embedding abd Various WSI-MIL architectures including global-attention, GCN, linear attention (TransMIL with absolute learnable embedding) and self-attention (HIPT with absolute embedding). Our Flash Attention with extrapolation relative position embedding (2d-ALiBi) show strong performance.}
  \vspace*{-5 mm}
  \label{T3} 
  \end{center}   
  \end{table}
\subsection{Slide-level Survival Prediction}
\noindent \textbf{Evaluation Metrics.}
For all the experiments, C-Index scores are reported for the 2 datasets. We follow the data splits and pretrained patch embedding proposed in HIPT\cite{chen2022scaling} for fair comparison. Also the performance results are reported via the mean and standard variance values of performance metrics by multiple folder cross-validation with the same running setting to HIPT \cite{chen2022scaling}. 

\noindent \textbf{Comparison with baselines.}
For this task, we use the survival cross-entropy loss proposed by Zadeh et al. \cite{zadeh2020bias}. The results are summarized in Table \ref{T3}, where we directly evaluate several survival prediction WSI-MIL methods, including AB-MIL\cite{pmlr-v80-ilse18a}, AMISL \cite{yao2020whole}, DS-MIL\cite{li2021dual}, GCN-MIL \cite{li2018graph}. Then we compare our method with some state-of-the-art combining position embedding on Transformer: TransMIL\cite{NEURIPS2021_10c272d0}  and HIPT\cite{chen2022scaling}. Though our method show some improvement, the C-index score is still too low to daily clinical usage depending on only WSI information. In the near future, we would like to investigate more on this task, e.g. combining multi-modality features as used in \cite{Chen_2021_ICCV,jaume2023modeling}, since Transformer also born with great ability on multi-modality fusion \cite{lee2018stacked,tsai2019multimodal,Chen_2021_ICCV,radford2021learning}.  

\subsection{Further Ablation Experiments}
\begin{figure}[htbp]
  \centering
   \begin{subfigure}{0.48\textwidth}
        \centering
        \includegraphics[width=0.9\textwidth]{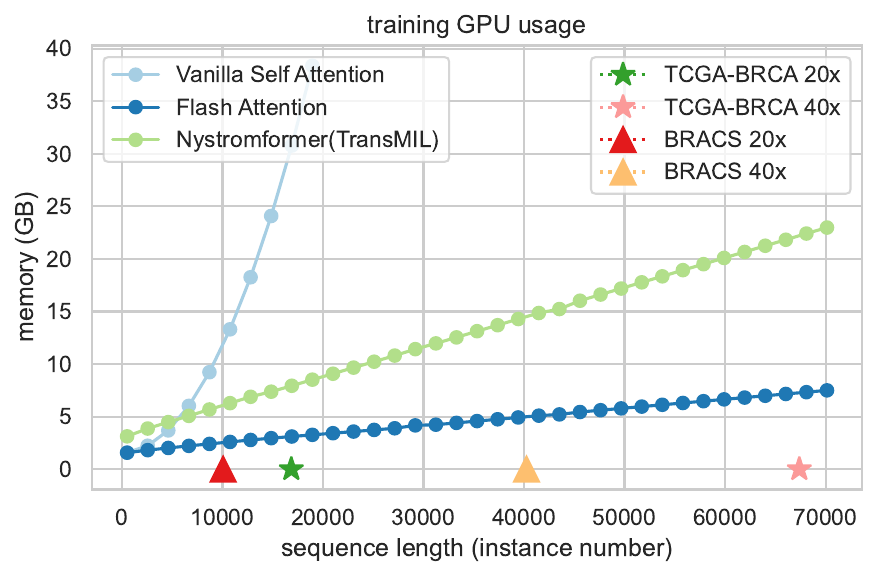}
    \end{subfigure}
    \hfill
    \begin{subfigure}{0.48\textwidth}
        \centering
        \includegraphics[width=0.9\textwidth]{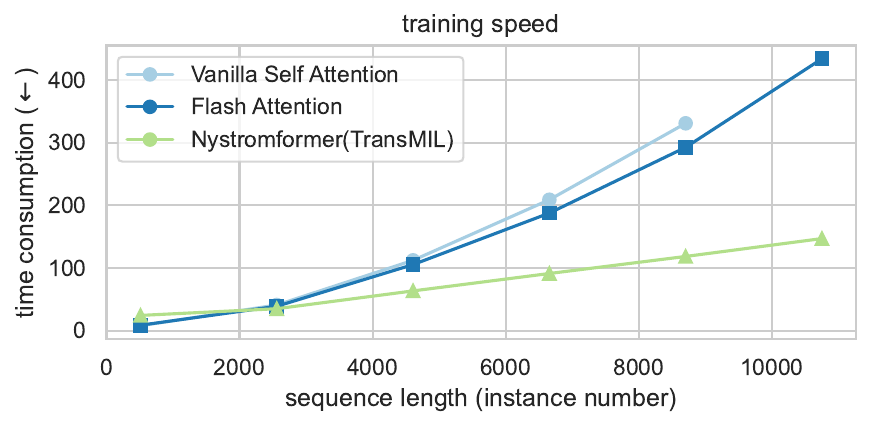}
    \end{subfigure}
   \caption{Training memory usage and speed using different Attentions. We also show markers about the max instance number of WSI used in this paper to show potentials on future higher magnitude learning.}
   \label{figure4}
   \vspace*{-5 mm}
\end{figure}

Here we provide ablations on the \textbf{training efficiency} of different transformer implementation as shown in Figure \ref{figure4}. We also provide some other performance ablations (number of Transformer blocks and multi-head, bias slope coefficient, weight decay, dropout ratio) in supplementary materials since Transformer is easy to be over-fitting on this task.

\section{Conclusion}
In this work, our proposed Long-contextual MIL (Long-MIL) method addresses the challenges in histopathology image analysis, offering superior performance in handling shape-varying Whole Slide Images (WSIs). By introducing Linear Bias into Attention and leveraging the Flash-Attention module, our approach enhances position embedding and tackles computational complexity, respectively. Extensive evaluations across four datasets affirm the effectiveness of Long-MIL in WSI classification and survival prediction tasks. Given the strong ability of long sequence modelling of our method, in the future we would like to adapt it longer sequence with higher resolution thus stronger information. Also, we will try to make more efforts on the unresolved problem of multi-modality survival prediction for more life-saving. 

{
    \small
    \bibliographystyle{ieeenat_fullname}
    \bibliography{main}
}


\end{document}